\documentclass[letterpaper, 10 pt, conference]{ieeeconf}  

\IEEEoverridecommandlockouts                              

\usepackage{times}
\usepackage{cite}
\usepackage{multicol}
\usepackage{graphicx}
\usepackage{caption}
\usepackage{subcaption}
\usepackage{amsfonts}
\usepackage{amsmath}
\usepackage{amssymb}
\usepackage{hyperref}
\usepackage{tabularx}
\usepackage{multirow}
\let\labelindent\relax
\usepackage{enumitem}
\usepackage{booktabs} 
\usepackage[symbol]{footmisc}

\usepackage[dvipsnames]{xcolor}

\title{\LARGE \bf
Affordance-Guided Reinforcement Learning via Visual Prompting
}

\author{Olivia Y. Lee$^{1}$, Annie Xie$^{1}$, Kuan Fang$^{2}$, Karl Pertsch$^{1, 3}$, Chelsea Finn$^{1}$ 
\thanks{$^{1}$Stanford University, $^{2}$Cornell University, $^{3}$University of California, Berkeley. Correspondence to: \href{mailto:oliviayl@stanford.edu}{\texttt{oliviayl@stanford.edu}}}
}

\begin{document}

\maketitle
\thispagestyle{empty}
\pagestyle{empty}

\begin{abstract}

Robots equipped with reinforcement learning (RL) have the potential to learn a wide range of skills solely from a reward signal. However, obtaining a robust and dense reward signal for general manipulation tasks remains a challenge. Existing learning-based approaches require significant data, such as human demonstrations of success and failure, to learn task-specific reward functions. Recently, there is also a growing adoption of large multi-modal foundation models for robotics that can perform visual reasoning in physical contexts and generate coarse robot motions for manipulation tasks. Motivated by this range of capability, in this work, we present Keypoint-based Affordance Guidance for Improvements (KAGI), a method leveraging rewards shaped by vision-language models (VLMs) for autonomous RL. State-of-the-art VLMs have demonstrated impressive zero-shot reasoning about affordances through keypoints, and we use these to define dense rewards that guide autonomous robotic learning. On diverse real-world manipulation tasks specified by natural language descriptions, KAGI improves the sample efficiency of autonomous RL and enables successful task completion in 30K online fine-tuning steps. Additionally, we demonstrate the robustness of KAGI to reductions in the number of in-domain demonstrations used for pre-training, reaching similar performance in 45K online fine-tuning steps.\footnote[2]{Project page: \href{https://sites.google.com/view/affordance-guided-rl}{\fontsize{7.5}{8} \texttt{sites.google.com/view/affordance-guided-rl}}}

\end{abstract}

\section{Introduction}

Developing generalist robots that can learn and adapt to new tasks autonomously without extensive human efforts has been a longstanding goal in robotics. Typically, this is achieved by collecting a large amount of demonstration data, often in the same domain and manually teleoperated by a human expert, which is costly and limits generalization. Ideally, robotic systems should allow human users to specify reward functions easily and flexibly with high-level instructions. Reinforcement learning (RL) holds great promise of learning new tasks autonomously by optimizing a reward signal. Prior work has sought to improve the sample efficiency of these algorithms by pre-training on large offline datasets~\cite{kumar2020conservative,ashvin2020accelerating,kostrikov2021offline,kumar2022pre,mark2022fine,yang2023robofume}. However, learning robust reward functions still requires careful engineering or large amounts of data~\cite{ho2016generative,fu2017learning,fu2018variational,xie2018few,singh2019end}.

Recent advances in large language models (LLMs) and vision-language models (VLMs) trained on Internet-scale data show promising results in breaking down complex language instructions into task plans~\cite{driess2023palm,brohan2023rt,saycan2022arxiv,chen2023open,huang2022language,huang2022inner}, performing visual reasoning in varied contexts~\cite{yang2023set,chen2024spatialvlm,nasiriany2024pivot}, and generating coarse robot motions for simple manipulation tasks~\cite{brohan2023rt,wang2023prompt,huang2022language,jiang2022vima,nasiriany2024pivot}. However, fine-tuning large pre-trained models to perform such tasks typically requires extensive human supervision, such as teleoperated demonstrations or feedback on whether the task was successfully completed.

A group of prior work uses the generalization capabilities of LLMs/VLMs to solve open sets of tasks through prompt engineering and in-context learning~\cite{liang2023code,huang2023voxposer}, but these works use external motion planners and skill primitives to generate low-level actions, which require significant engineering efforts and are not robust. Another line of work prompts LLMs/VLMs to generate rewards via code to train policies or plan motions to solve the task~\cite{ma2023eureka,yu2023language,shah2022lmnav}. However, existing approaches are only applied to high-level state spaces, navigation, simulated tasks, or coarse manipulation tasks, and can fall short on real-world manipulation tasks that require sophisticated spatial reasoning over visual inputs.

\begin{figure}
    \centering
    \vspace*{0.1cm}
    \includegraphics[trim={0cm 0cm 0cm 0cm},clip,width=\linewidth]{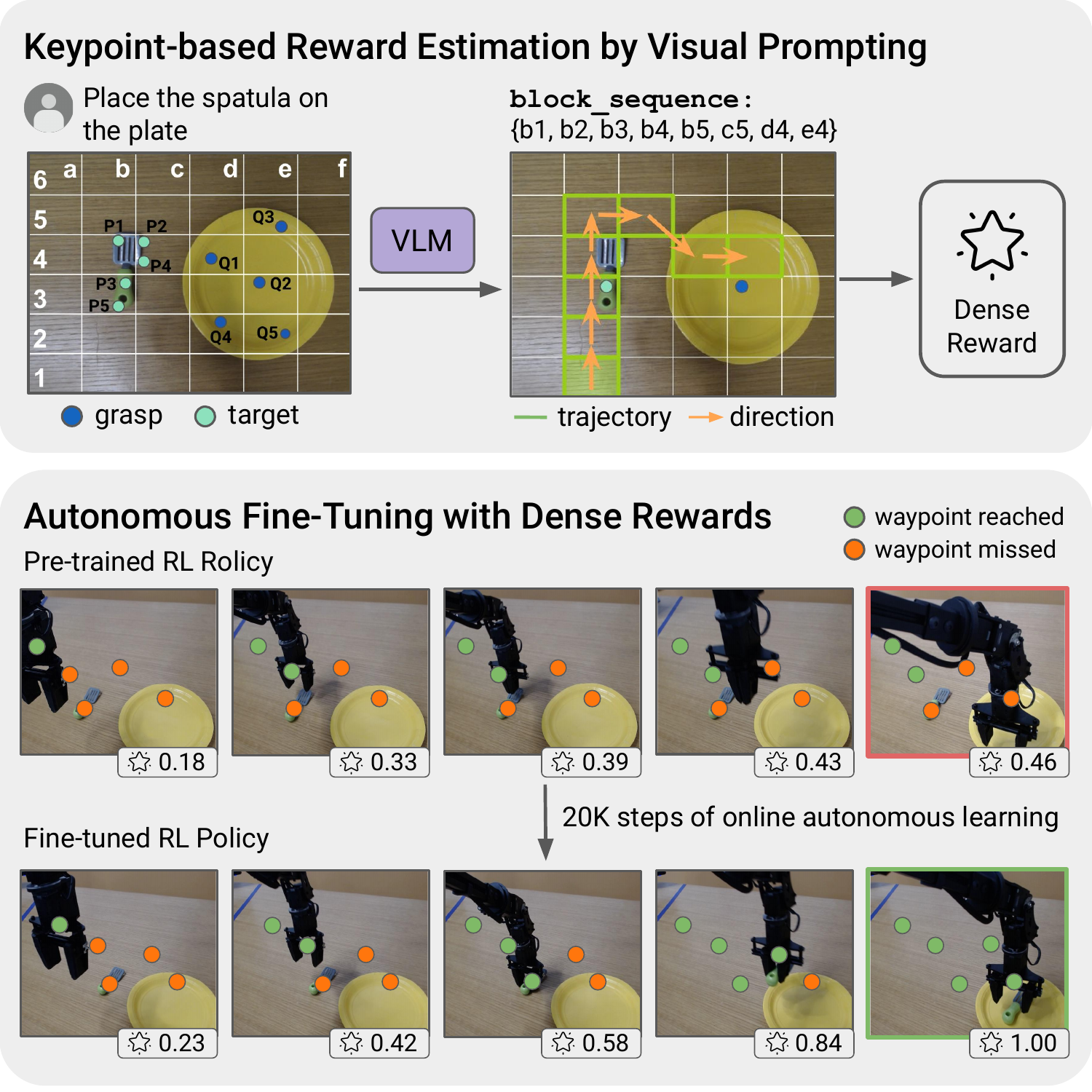}
    \caption{\small \textbf{Keypoint-based Affordance Guidance for Improvements (KAGI)} computes dense rewards defined using affordance-based keypoints and waypoints trajectories inferred by a VLM. Our dense reward formulation helps to shape learned behaviors, facilitating efficient online fine-tuning across diverse real-world tasks.
    }
    \label{fig:teaser}
    \vspace*{-1cm}
    \hspace*{0.05cm}
\end{figure}

Defining explicit rewards using pre-trained VLMs is hence an attractive approach for learning manipulation tasks. Thus far, VLMs have mainly been used to generate sparse rewards~\cite{mahmoudieh2022zeroshotreward,villa2023pivot,yang2023robofume}, often leading to less efficient learning. VLMs hold much richer geometric and semantic representations that we can elicit, such as reasoning about the affordances of objects and environments. In this work, we leverage insights on effective visual prompting techniques \cite{yang2023set} to generate affordance keypoints and waypoint trajectories, from which we directly derive dense rewards to improve the efficiency of online fine-tuning with RL. 

We present \textbf{K}eypoint-based \textbf{A}ffordance \textbf{G}uidance for \textbf{I}mprovements (KAGI), a method for open-vocabulary visual prompting to extract rewards from VLMs for autonomous RL. Our method extracts affordance keypoint representations zero-shot from VLMs and computes waypoint-based dense rewards (Fig.~\ref{fig:teaser}), which we integrate into an autonomous RL system that uses sparse rewards from a fine-tuned VLM~\cite{yang2023robofume}. KAGI demonstrates improved success rates on complex manipulation tasks while being robust to reduced in-domain expert demonstrations, which systems using only sparse rewards or open-loop primitives struggle to generalize to.

\section{Related Work} \label{sec:literature}


\subsection{Foundation Models for Robotics}
\textbf{Planning and high-level task reasoning.}
The rapid development of foundation models has drawn significant attention~\cite{bommasani2022opportunities}, demonstrating that models trained on Internet-scale data are highly adaptable to a wide range of downstream tasks, including robotics. Foundation models have enabled high-level planning and reasoning for embodied agents \cite{huang2022language,zeng2022socraticmodels,saycan2022arxiv,huang2022inner,raman:neurips22fmdm,Silver2023GeneralizedPI,LinAgiaEtAl2023}, generated sub-goals \cite{dorbala2022clipnavusingclipzeroshot, shah2022lmnav, huang23vlmaps}, and generated coarse motions for manipulation \cite{rt22023arxiv, huang2022language, nasiriany2024pivot,wang2023prompt,jiang2022vima}. These impressive results show great potential in harnessing the rich knowledge and reasoning capabilities of foundation models for tackling robotic manipulation tasks. 

Despite the rapid development of LLMs for robotic control, recent works have investigated using multi-modal models to jointly provide visual and language prompts to robotic agents \cite{driess2023palm,nasiriany2024pivot,jiang2022vima,gu2023rttrajectory}. These approaches recognize that converting visual observations into language descriptions and planning solely with language loses rich information from the visual modality for scene understanding. Though LLMs can provide general priors for spatial planning and reasoning, they must be properly grounded to generate accurate responses about real-world environments~\cite{3dllm,huang2023chat,wang2023chat,huang2023voxposer}. Our work harnesses state-of-the-art VLMs to facilitate reasoning in both language and image domains. We leverage the image domain by computing rewards in image space to bridge high-level LLM plans with low-level actions for embodied tasks.  

\textbf{Spatial reasoning with VLMs.}
Language provides a highly natural interface for providing task instructions and specifying goals. That said, robotics relies heavily on accurately perceiving and interacting with the environment. Modern VLMs, such as GPT-4o [\citenum{openai2024gpt4ocard}] and Gemini [\citenum{geminiteam2024gemini}], demonstrate promising capabilities combining reasoning with environment perception via visual inputs. Recent approaches have explored using VLMs for shaping rewards via code~\cite{ma2023eureka,yu2023language}, reward estimation~\cite{nair2022r3m,shah2022lmnav}, or preference-based learning~\cite{wang2024rlvlmf}. Our work focuses on directly obtaining rewards \textit{zero-shot} from VLMs by leveraging the rich geometric and semantic representations that VLMs trained on Internet-scale data possess, such that VLM-generated dense rewards can be used effectively in closed-loop autonomous RL systems. Our approach is inspired by works that use generalizable and accurate keypoint-based reasoning of VLMs for zero-shot manipulation~\cite{huang2024rekep,nasiriany2024pivot,yang2023set}, translating high-level plans into low-level robot actions. Beyond zero-shot open-loop manipulation, our work presents a novel application of open-vocabulary visual prompting, which enables us to improve policies pre-trained on diverse data via online RL fine-tuning with VLM-defined dense rewards.

Modern VLMs can derive rewards in image space for learning state-action mappings via RL, enabling robots to learn through trial-and-error without training skill policies via imitation learning which are costly and difficult to scale. A key contribution of our work is using VLMs to determine task completion (sparse rewards) and specify intermediate waypoints as goals (dense rewards), where prior works focus mainly on the former~\cite{mahmoudieh2022zeroshotreward,villa2023pivot,yang2023robofume}. 
Our work explores using mark-based visual prompting (e.g., grid lines and object segmentations~\cite{ravi2024sam2}) to augment raw image observations and guide semantic reasoning in VLMs, leveraging pre-trained representations in VLMs as zero-shot reward predictors.


\subsection{Autonomous Reinforcement Learning}
Online RL is the paradigm by which an agent gathers data through interaction with the environment, then stores this experience in a replay buffer to update its policy. This contrasts with offline RL, where an agent updates its policy using previously collected data without itself interacting with the environment. A longstanding goal is autonomous RL, an agent that autonomously gathers real-world experience online. This approach holds great promise for scalable robot learning as agents learn through their own experience and do not require manual environment resets between trials \cite{sharma2022autonomous}.

Autonomous RL is difficult to implement in the real world, with the primary challenges being sample complexity, providing well-shaped rewards, and continual reset-free training. Several works have developed reset-free systems that reduce human interventions \cite{balsells2023gear,yang2023robofume, gupta2021resetfree,sun2022autonomous}, but reward shaping is an open problem. As manually specified rewards are difficult to engineer and easy to exploit, there is great potential to learn rewards from offline data or large pre-trained models. The large bank of image and video datasets, plus the fast inference speed and accessibility of large pre-trained models, could provide more precise and informative shaping rewards.

RoboFuME~\cite{yang2023robofume} is a recent work leveraging a reset-free pre-train fine-tune paradigm to train manipulation policies via autonomous RL. RoboFuME pre-trains offline RL policies on diverse offline data~\cite{ebert2021bridge, walke2023bridgedata} and in-domain demonstrations, then fine-tunes a sparse reward classifier [\citenum{zhu2023minigpt4}] for online policy fine-tuning. However, online RL suffers when reward signals are too sparse, and fine-tuning the reward classifier requires a substantial number of in-domain demonstrations per task, which is costly but also makes the system more brittle and less robust to generalization. Our work similarly focuses on the pre-train fine-tune paradigm, specifically on improving the efficiency and robustness of online fine-tuning, enabling effective adaptation of RL policies pre-trained on diverse datasets. Inspired by recent work extracting rewards from VLMs to guide zero-shot robotic manipulation \cite{villa2023pivot, mahmoudieh2022zeroshotreward} and online adaptation \cite{yang2023robofume, xiong2024adaptive} , we leverage affordance representations extracted zero-shot from VLMs to tackle the dense reward shaping problem for online RL.

\begin{figure*}[htp!]
    \centering
    \vspace*{0.1cm}
    \includegraphics[trim={0.1cm 5.3cm 2cm 0.25cm},clip,width=0.95\textwidth]{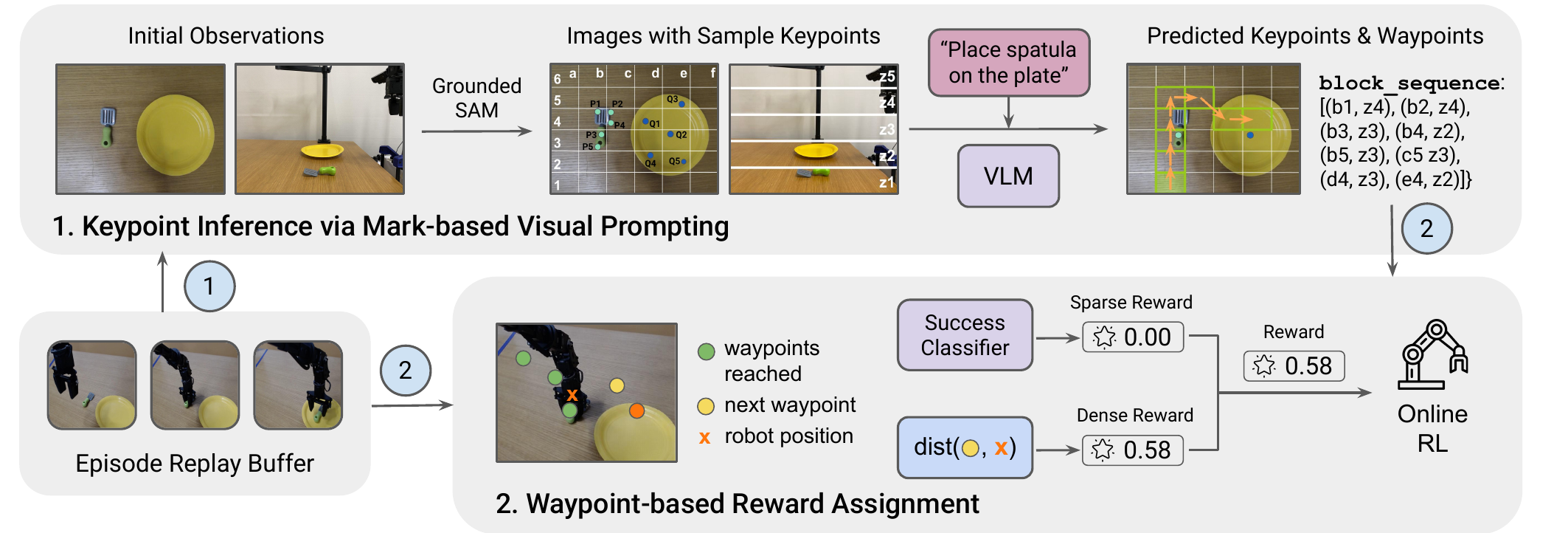}
    \vspace*{-0.05cm}
    \caption[Diagrammatic illustration of KAGI.]{\small \textbf{KAGI} consists of two components. The first, represented by arrow labeled 1 above, leverages a VLM to select from a set of affordance keypoints, then generate a waypoint sequence. The second, represented by arrows labeled 2 above, involves per timestep reward computation for each frame in the episode replay buffer, computing dense reward with respect to the waypoint sequence and a sparse reward derived from a success classifier. The dense reward is used for online RL if the sparse reward is 0, else the sparse reward is used.}
    \label{fig:method}
    \vspace*{-0.45cm}
\end{figure*}

\section{Keypoint-based Affordance Guidance for Improvements (KAGI)} 
\label{sec:methodology}

\subsection{Problem Statement} We propose Keypoint-based Affordance Guidance for Improvements (KAGI) to facilitate autonomous fine-tuning on unseen tasks as shown in Fig.~\ref{fig:method}. We consider problems that can be formulated as a partially observable Markov Decision Process (POMDP) tuple $(\mathcal{S}, \mathcal{A}, \mathcal{O}, \gamma, f, p, r, d_0)$ where $\mathcal{S}$ is the state space, $\mathcal{A}$ is the action space, $\mathcal{O}$ is the observation space, $\gamma \in (0, 1)$ is the discount factor, $r(s, a)$ is the reward function and $d_0(s)$ is the initial state distribution. The dynamics are governed by a transition function $p(s'|s, a)$. The observations are generated by an observation function $f(o | s)$. The goal of RL is to maximize the expected sum of discounted rewards $\mathbb{E}_\pi [\sum_{t=1}^{\infty} \gamma^t r(s_t, a_t) ]$.
In this work, we use RGB image-based observations.
The reward function is typically hand-engineered or learned, for instance via examples of task success and failure~\cite{ho2016generative,fu2017learning,fu2018variational,xie2018few,singh2019end,yang2023robofume}.
We assume the existence of a sparse task completion reward (i.e., $r(s, a) \in \{ 0, 1 \}$), acquired with systems like RoboFuME~\cite{yang2023robofume}. 

Our system consists of an offline pre-training phase and an online fine-tuning phase, where the latter phase requires a reward function to label successes and failures. A sparse reward is typically easier to specify but, with it, RL algorithms require more samples to learn a successful policy, because the agent must encounter success through its own exploration. In comparison, a dense reward provides continuous feedback that guides the agent towards success. KAGI aims to provide the latter type of feedback by augmenting sparse task completion rewards with dense shaping rewards. Specifically, dense rewards are calculated with respect to intermediate waypoints marking trajectory points towards the goal. We find that such guidance can facilitate more efficient and generalizable online learning than sparse rewards alone.

\subsection{Keypoint-Based Reward Estimation via Visual Prompting} \label{sec:our-approach} 

We employ an vision language model (VLM) to estimate dense rewards to facilitate fine-tuning of a pre-trained policy.
First, we prompt the VLM to select appropriate grasp and target keypoints for the task, inspired by recent visual prompting techniques.
Then, the estimated rewards are used to generate a coarse trajectory of how the robot should complete the task. We then assign rewards to each timestep of an RL episode based on how well it follows this trajectory. 

\begin{figure} 
    \centering
    \vspace*{0.2cm}
    \begin{subfigure}{0.32\linewidth}
        \centering
        \includegraphics[width=\textwidth]{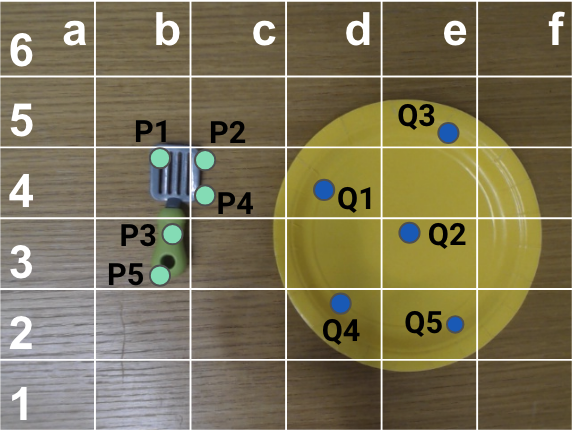}
        \caption{\small Top-down view}
        \label{fig:top-down-ex}
    \end{subfigure}
    \begin{subfigure}{0.32\linewidth}
        \centering
        \includegraphics[width=\textwidth]{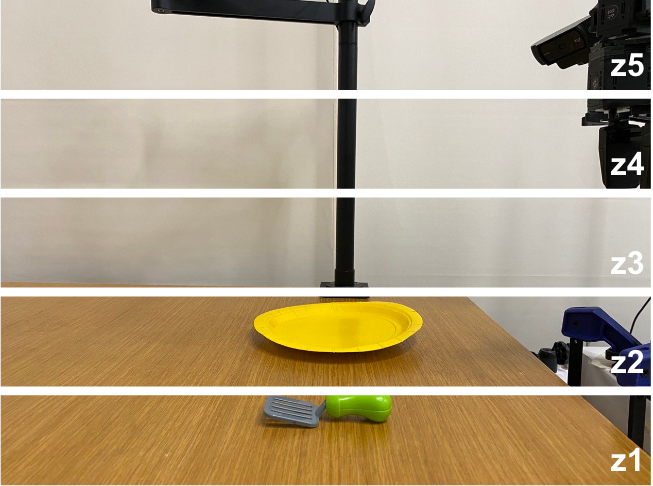}
        \caption{\small Side view}
        \label{fig:side-ex}
    \end{subfigure}
    \begin{subfigure}{0.32\linewidth}
        \centering
        \includegraphics[width=\textwidth]{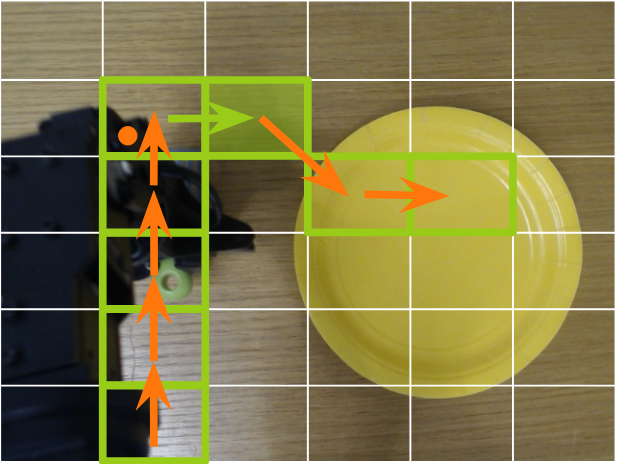}
        \caption{\small VLM Trajectory}
        \label{fig:closest-block-ex}
    \end{subfigure}
    \vspace*{-0.1cm}
    \caption{\small Example of annotated images to VLM (\ref{fig:top-down-ex}, \ref{fig:side-ex}) and VLM-generated trajectory (\ref{fig:closest-block-ex}). In \ref{fig:top-down-ex}, teal points labeled P1-5 denote grasp keypoints, blue points labeled Q1-5 denote target keypoints for the VLM to select from. In \ref{fig:closest-block-ex}, orange point is robot position, green tiles denote the generated trajectory, and arrows denote motion direction. KAGI's dense reward formulation encourages the robot to move to the next block, following the green arrow.}
    \vspace*{-0.5cm}
\end{figure}

We take the first observation in the episode $o_0$, consisting of a top-down image $o_0^{d}$ and a side view image $o_0^s$. Our goal is to create a candidate set of grasp and target keypoints that GPT-4o can select from. We preprocess these images by (1) passing $o_0^d$ through Grounded SAM~\cite{ren2024groundingsam} to get segmentations of relevant objects and sample five points from their masks via farthest point sampling from the centroid, and (2) overlaying a $6\times6$ grid and the sampled keypoints on top of $o_0^d$ to get $\tilde{o}^d$. For the side-view image, we augment $o_0^s$ with evenly-spaced labeled horizontal lines to get $\tilde{o}^s$, providing depth information that is not available in the top-down view. See Figures \ref{fig:top-down-ex} and \ref{fig:side-ex} for sample processed inputs.
We use this density of grid lines as preliminary tests found that this was most suitable for GPT-4o to reason with. We pass $\tilde{o}^d$ and $\tilde{o}^s$ with a language instruction and metaprompt to GPT-4o, which generates \texttt{block\_sequence}. This sequence is a list of tuples $(x,y,z)$, where $(x, y)$ is a grid point chosen from $\tilde{o}^d$ representing a position in the xy-plane from top-down and $z$ is chosen from $\tilde{o}^s$ representing height in the z-axis from the side view. Dense rewards are calculated with respect to \texttt{block\_sequence} in the next step.

For each frame in the episode, we compute the robot position in image space. We use two RANSAC regressors fitted on 50 manually collected keypoints, one for each camera view, to compute the robot position $(x_\text{rob}^t, y_\text{rob}^t, z_\text{rob}^t)$, where $(x_\text{rob}^t, y_\text{rob}^t)$ is from the top-down and $z_\text{rob}^t$ is from the side view. Optionally, we can use pixel trackers [\citenum{karaev2023cotracker}] to track a specific point on the object $(x_\text{obj}^t, y_\text{obj}^t, z_\text{obj}^t)$ to additionally define rewards based on object poses. Using these coordinates, we compute the nearest block to the robot position in \texttt{block\_sequence}, $B_\text{rob}^i$ (see Figure \ref{fig:closest-block-ex} for an example). We compute a reward $r_\textit{rob}^t$ is based on the L2 distance from the robot position to the \textit{block after the closest block in} \texttt{block\_sequence}, $B_\text{rob}^{i+1}$, to encourage progression towards the goal and avoid stagnating at the current position. We transform the distance such that reward is between $0$ and $1$. The reward is $1$ when the sparse reward is $1$, else it is our dense reward. This reward is optimized using an online RL algorithm~\cite{sharma2023medal++}.

\subsection{System Summary} \label{sec:implement}
To optimize this reward, we build on RoboFuME~\cite{yang2023robofume}, an autonomous RL pipeline that learns from image observations. RoboFuME requires a set of forward and backward task demonstrations, and pre-trains a policy on this data and the Bridge dataset~\cite{ebert2021bridge, walke2023bridgedata}. For the reward model, it fine-tunes MiniGPT-4~\cite{zhu2023minigpt4} as a success classifier using the in-domain demonstrations and a few additional failure examples. 
We augment these sparse rewards with our waypoint-based dense rewards to combine the benefits of both modes of feedback.


From the VLM-generated waypoint trajectory, we use the centroid of each grid tile and height from the side-view to create a 3D coordinate trajectory. We calculate the L2 distance between the current robot position and target waypoint in image space, and pass each distance through a modified tanh function, so $r_\text{dense} = 0.5(1 - \tanh{(\lambda(d_t - \varphi))}$, where $d_t$ is the L2 distance between the robot position and target waypoint at timestep $t$. The scaling factor $\lambda$ and offset $\varphi$ are hyperparameters; $\lambda = 0.1, \text{ } \varphi = 15$ for the simulation experiments in Section \ref{sec:sim-expts}, and $\lambda = 0.02, \text{ } \varphi = 100$ for the real-robot experiments in Section \ref{sec:fine-tuning}. This ensures the dense reward stays between 0 and 1, with values closer to 1 when the robot trajectory is closer to the VLM-generated waypoint trajectory. What distinguishes trajectories that stay close to the VLM-generated trajectories and truly successful trajectories is the sparse reward. We set the final reward for each timestep $r = r_\text{sparse}$ if $r_\text{sparse} = 1$ else $r = r_\text{dense}$.

\section{Experiments} 
Our experiments aim to answer the following questions:
\begin{itemize}
\item Does our VLM-based dense reward formulation improve the efficiency of online RL? (Sec.~\ref{sec:fine-tuning})
\item Is our autonomous RL system robust to fewer in-domain demonstrations? (Sec.~\ref{sec:reduce})
\end{itemize}
We study tasks in both simulation and on a real robot to answer these questions, which we describe next.


\subsection{Experimental Setup}
\begin{figure}
    \centering
    \vspace*{0.2cm}
    \includegraphics[trim={0.05cm 0cm 0cm 0cm},clip,width=\linewidth]{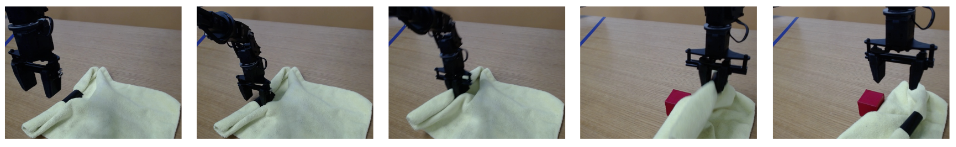} \\ %
    \vspace*{0.1cm}
    
    \includegraphics[trim={0.1cm 0cm 0.1cm 0cm},clip,width=\linewidth]{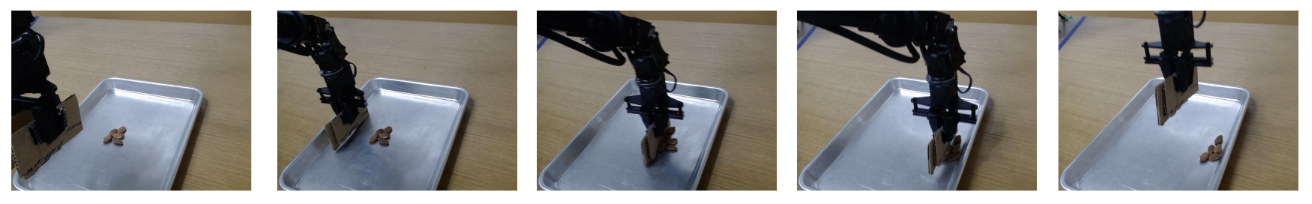} \\ %
    \vspace*{0.1cm}
    \includegraphics[width=\linewidth]{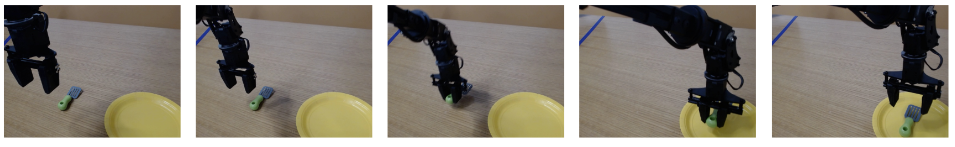} \\ %
    \vspace*{0.1cm}
    \includegraphics[trim={0.01cm 0cm 0cm 0cm},clip,width=\linewidth]{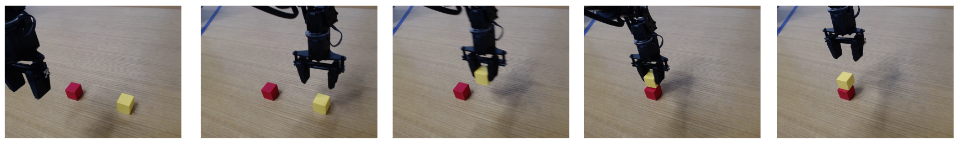}
    \caption{
    \small \textbf{Tasks Visualization.} Evaluation is conducted on four real-world tasks: Cloth Covering (Deformables Manipulation), Almond Sweeping (Non-Prehensile Manipulation), Spatula Pick-Place (Functional Grasping), and Cube Stacking (Precise Manipulation).}
    \label{fig:real-robot-tasks}
    \vspace*{-0.5cm}
\end{figure}

\textbf{Tasks.} In PyBullet simulation, we study a \emph{Bin-Sorting} task~\cite{kumar2022pre,yang2023robofume}, where a WidowX 250 arm needs to sort objects into the correct bin specified by the language instruction. On a physical WidowX 250 robot arm, we study \emph{Cloth Covering}, \emph{Almond Sweeping}, \emph{Spatula Pick-Place}, and \emph{Cube Stacking} (visualized in Fig.~\ref{fig:real-robot-tasks}). Our tasks span a range of complex skills involving manipulation of deformables, non-prehensile manipulation, functional grasping of objects handles, and precise grasping and placement of $3$cm small cubes designed to challenge our system. Each of our tasks consists of a forward component (e.g., uncovering the block from under the cloth) and backward component (e.g., covering the block with the cloth), and therefore can be autonomously practiced by alternating between these two task components.

\textbf{Comparisons.} To assess performance without any online fine-tuning, we compare KAGI to language-conditioned behavior cloning and a pre-trained RL baseline, \emph{CalQL}~\cite{nakamoto2023calql}.
We also evaluate \emph{RoboFuME}~\cite{yang2023robofume}, which fine-tunes the pre-trained RL policy online with sparse rewards derived from a VLM success classifier. This comparison allows us to understand the benefits of the proposed dense rewards. Finally, we evaluate \emph{MOKA}~\cite{liu2024moka}, which executes the trajectory defined by the VLM-generated waypoints. This last comparison does not perform any additional learning or fine-tuning.



\textbf{Datasets.}
All methods except for MOKA are pre-trained on a combination of trajectories selected from the Bridge dataset~\cite{ebert2021bridge,walke2023bridgedata} and a set of in-domain demonstrations. The in-domain demonstrations consist of $50$ forward trajectories and $50$ backward trajectories. Following RoboFuME~\cite{yang2023robofume}, we additionally collect $20$ failure trajectories, which are used to train the VLM-based success classifier used by RoboFuME and KAGI. In Section \ref{sec:reduce}, we test these methods on a $5\times$ reduction in the quantity of in-domain demonstrations.

\textbf{Evaluation.} In simulation and real-world evaluations, we roll out each forward task policy for $20$ trials. For the real-world tasks, success is evaluated as follows. For Cube Covering, the entire cube must be uncovered from the camera perspective shown in Fig.~\ref{fig:real-robot-tasks}; for Almond Sweeping, all five almonds must reach the right side of the tray; for Spatula Pick-Place, the spatula must be on the plate; for Cube Stacking, the yellow cube must be stable atop the red cube. 


\begin{table}[]
    \centering
    \vspace*{0.2cm}
    \begin{tabular}{llllll}
    \toprule
    Task & BC \hspace{0.2cm} & CalQL & RoboFuME & KAGI (Ours) \\
    \midrule
    Cloth Covering & 35\% & 55\% & \textbf{80\%} & \textbf{80\%} \\
    Almond Sweeping & 25\% & 45\% & 70\% & \textbf{80\%} \\
    Spatula Pick-Place & 30\% & 25\% & 45\% & \textbf{65\%} \\
    Cube Stacking & 10\% & 15\% & 25\% & \textbf{45\%} \\ \bottomrule
    \end{tabular}
    \caption{\small
    Success rates over $20$ trials for each method on four real-world tasks. We compare KAGI (fine-tuned for $30$K steps) to offline-only methods and RoboFuME (fine-tuned for $30$K steps).
    }
    \label{tab:dense-shaping}
    \vspace*{-0.5cm}
\end{table} 

\subsection{Simulation Experiments} \label{sec:sim-expts}
We conduct a preliminary test of our dense reward formulation in a simulated Bin-Sorting task.
Following the experimental setup in RoboFuME~\cite{yang2023robofume}, each approach is provided $10$ forward and backward demonstrations, $30$ failure demonstrations, and $200$ total demonstrations of other pick-place tasks of diverse objects.
In simulation, rewards are computed in $100 \times 100$ pixel image space. We compare three reward formulations: dense waypoint-based only, sparse success-based only (RoboFuME), and dense + sparse (KAGI).

\begin{figure}[htp!]
    \centering
    \begin{subfigure}{0.165\textwidth}
      \centering
      \includegraphics[width=\textwidth]{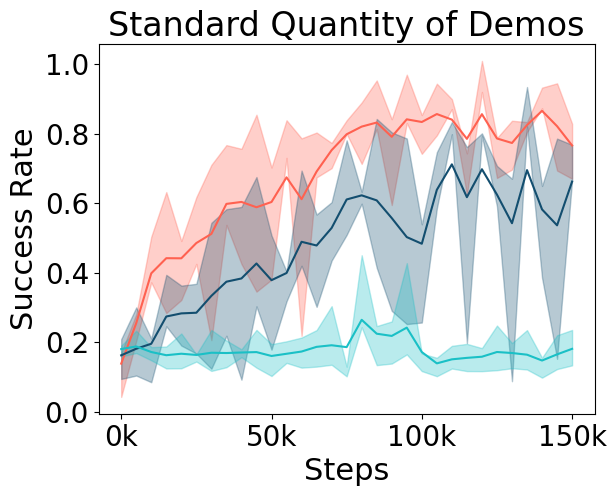}
      \vspace*{-0.2cm}
    \end{subfigure}%
    \begin{subfigure}{0.165\textwidth}
      \centering
      \includegraphics[width=\textwidth]{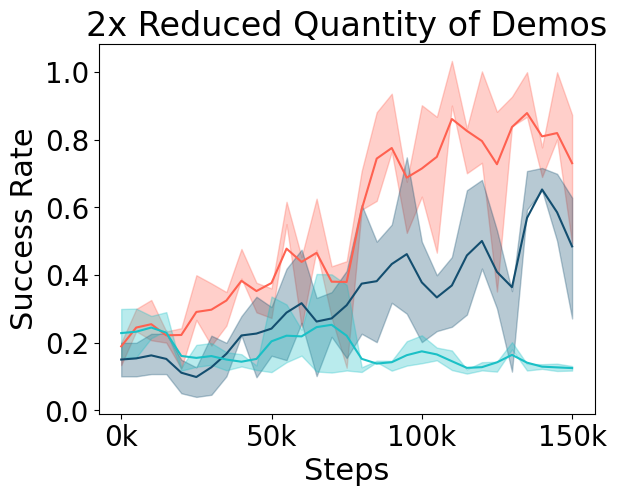}
      \vspace*{-0.2cm}
    \end{subfigure}%
    \begin{subfigure}{0.165\textwidth}
      \centering
      \includegraphics[width=\textwidth]{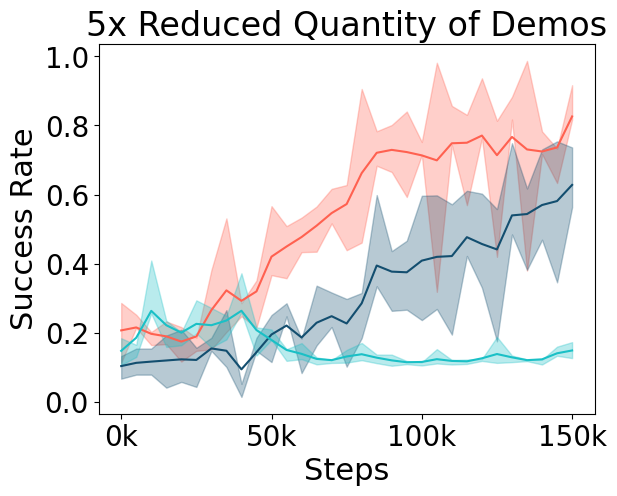}
      \vspace*{-0.2cm}
    \end{subfigure}
    \begin{subfigure}{0.5\textwidth}
      \centering
      \includegraphics[trim={0 0 0 12.25cm},clip,width=\textwidth]{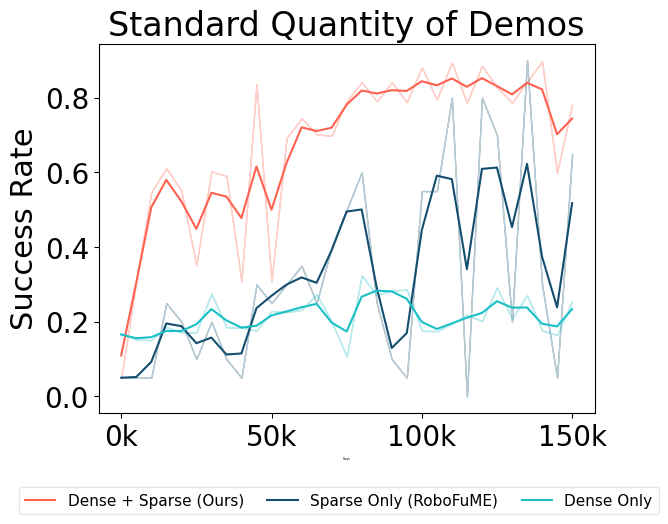}
    \end{subfigure}%
    \caption{\small Average success across $3$ seeds on simulated Bin-Sorting. We evaluate each reward formulation under: the standard number of demos (left), $2\times$ reduction (middle), and $5\times$ reduction (right).}
    \label{fig:sim-results}
\end{figure}

As shown in Fig.~\ref{fig:sim-results}, \textbf{our reward formulation achieves higher success rates than the sparse-only reward formulation}. The dense-only reward formulation performs worse than the other two. This indicates that both types of rewards are necessary for efficient and successful learning: while dense rewards are useful in shaping learned behaviors, sparse rewards are also crucial to distinguish truly successful trajectories. We further test our reward formulation succeeds with a reduced number of demonstrations of the task. We test the effect of a $2\times$ reduction in demonstrations ($10$ forward and $10$ backward demonstrations) and a $5\times$ reduction in demonstrations ($4$ forward and $4$ backward demostrations).
In Fig.~\ref{fig:sim-results}, we see that \textbf{our reward formulation is robust to these reductions in quantity of task demonstrations}.

\subsection{Real-World Fine-Tuning with Dense Rewards} \label{sec:fine-tuning}
In our real-world experiments, we evaluate RoboFuME and KAGI after fine-tuning for $30$K online environment training steps, with rewards computed in $640 \times 480$ pixel image space.
The results are shown in Table~\ref{tab:dense-shaping}.
For each task, RoboFuME fine-tuning improves over CalQL policies by $10$-$25$\%, which is consistent with the performance in~\cite{yang2023robofume} and confirms that further online fine-tuning is beneficial.
Notably, we see lower success rates on Cube Stacking, which requires high precision and is considerably harder than all the pick-place tasks in RoboFuME~\cite{yang2023robofume}.
With KAGI, task performance is similar or further increases across the board.

\begin{figure} 
    \centering
    \vspace*{0.2cm}
      \includegraphics[width=0.15\linewidth]{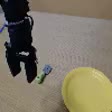}
      \includegraphics[width=0.15\linewidth]{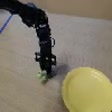}
      \includegraphics[width=0.15\linewidth]{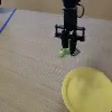}
      \includegraphics[width=0.15\linewidth]{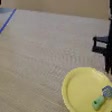}
      \includegraphics[width=0.15\linewidth]{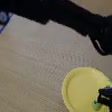}
      \includegraphics[width=0.15\linewidth]{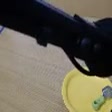} \\
      \vspace{0.3cm}
      \includegraphics[width=0.15\linewidth]{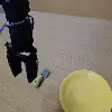}
      \includegraphics[width=0.15\linewidth]{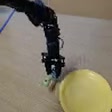}
      \includegraphics[width=0.15\linewidth]{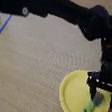}
      \includegraphics[width=0.15\linewidth]{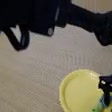}
      \includegraphics[width=0.15\linewidth]{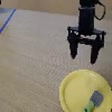}
      \includegraphics[width=0.15\linewidth]{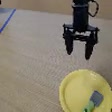}
    \caption{\small
    Qualitative examples of policies fine-tuned with RoboFuME (top) and KAGI (bottom). RoboFuME drops the spatula from a higher height and unnecessarily moves right after. KAGI places the spatula down more gently and moves to a neutral position after.
    }
    \label{fig:fine-tuned-behavior}
\end{figure}

While the increase in success rates 
achieved by KAGI over RoboFuME
is modest using the standard number of in-domain demosntrations for pre-training, 
we notice qualitatively different behaviors learned by each policy.
As an example, Fig.~\ref{fig:fine-tuned-behavior} shows a representative trajectory demonstrating policy performance for Spatula Pick-Place. The policy fine-tuned with RoboFuME, while fairly successful, commonly demonstrated the behavior
of dropping the spatula onto the plate from a height, rather than lowering and placing the spatula like the demonstrations. The robot gripper also continues moving rightward rather than staying above the placement point, indicating some coincidental successes. In comparison, KAGI's policy fine-tuned with dense shaping rewards showed behavior that matched the demonstrations more closely.
For more examples comparing qualitative behavior across tasks, please see our supplementary video.

\subsection{Robustness to Reduced In-Domain Demonstrations} \label{sec:reduce}
Both RoboFuME and KAGI pre-train on a set of $100$ in-domain demonstrations for each new task, which is a nontrivial cost.
We investigate the effect of reducing the number of in-domain demonstrations used by both methods.
We pre-train an offline RL policy via CalQL on $5\times$ fewer than the standard quantity of in-domain demonstrations ($10$ forward tasks, $10$ backward tasks, and $4$ failures). We then fine-tune this policy online for $45$K steps with RoboFuME and with KAGI. The results are shown in Table~\ref{tab:less-demos}.

\begin{table} 
    \centering
    \begin{tabular}{llll}
    \toprule
    Task & CalQL & RoboFuME & KAGI (Ours) \\
    \midrule
    Cloth Covering & 45\% & 50\% & \textbf{75\%} \\
    Almond Sweeping & 40\% & 55\% & \textbf{80\%} \\
    Spatula Pick-Place & 10\% & 30\% & \textbf{60\%} \\
    Cube Stacking & 5\% & 15\% & \textbf{40\%} \\
    \bottomrule
    \end{tabular}
    \caption{\small Success rates over $20$ trials for each method with $5\times$ fewer in-domain demonstrations.
    KAGI achieves similar performance to the setting with the standard amount of demonstrations, while  RoboFuME performance drops with fewer demonstrations.
    }
    \label{tab:less-demos}
    \vspace*{-0.5cm}
\end{table}

\textbf{Impact of reduced demos on the policy.}
Unsurprisingly, CalQL performs worse than when trained on the standard quantity of in-domain demonstrations.
Fine-tuning for $45$K steps using RoboFuME somewhat improves success rate over offline RL policies, but we see a significant gap with $5\times$ less demos compared to its original performance with the standard number of demonstrations. Therefore, the performance of RoboFuME depends heavily on the amount of in-domain demonstrations provided.
However, across all tasks, KAGI reaches close to its original performance even with this reduction. This small drop indicates that, \textbf{even with fewer in-domain demonstrations, KAGI can recover comparable performance with more fine-tuning}, as it uses dense shaping rewards to guide its behavior. Since KAGI can tolerate fewer task-specific demonstrations, it can be applied to new tasks more easily, with a smaller data burden.

\textbf{Impact of reduced demos on the success classifier.}
The reduction in in-domain data crucially impacts the behavior of the VLM-based (MiniGPT-4) success classifier used by both RoboFuME and KAGI.
We observed the sparse reward predictions of MiniGPT-4 were significantly worse, indicating fine-tuning the success predictor is reliant on substantial high-quality in-domain data to generate accurate sparse rewards.
To mitigate this issue and eliminate the confounding factor of inaccurate sparse rewards, we modify RoboFuME's~\cite{yang2023robofume} original sparse reward computation: the reward predictor was provided four task-completion prompts as input (e.g., `Is the spatula on the plate?', `Has the spatula been moved to the plate?', etc.), and must reach a consensus across all prompts to generate a sparse reward of $1$.
This reduces the number of false positives, which RL algorithms often exploit. The decreased accuracy of the sparse reward predictor is a key factor behind the minimal improvements of RoboFuME over offline RL, revealing another source of fragility in the system and underscoring the importance of dense shaping rewards for data-efficient fine-tuning.

\begin{table} 
    \centering
    \vspace*{0.25cm}
    \begin{tabular}{lccc}
    \toprule
    Task & \shortstack{MOKA with\\precise resets} & \shortstack{MOKA with\\imprecise resets} & KAGI (Ours) \\
    \midrule
    Cloth Covering & 90\% & 50\% & 80\% \\
    Almond Sweeping & 100\% & 65\% & 80\% \\
    Spatula Pick-Place & 70\% & 30\% & 65\% \\
    Cube Stacking & 60\% & 15\% & 45\% \\
    \bottomrule
    \end{tabular}
    \caption{\small
    Success rate of MOKA with precise resets of the objects and with slight perturbations to the object positions. Without precise resets, MOKA's performance drops significantly.
    }
    \label{tab:moka}
    \vspace*{-0.5cm}
\end{table}

\subsection{Importance of Online Fine-tuning}
Since the VLM prompting component of KAGI is inspired by MOKA~\cite{liu2024moka}, we additionally evaluate MOKA to understand whether online fine-tuning offers any advantages.
MOKA leverages GPT-4V to predict affordance keypoints and plan a sequence of action primitives (e.g., `lift', `reach grasp', `grasp'), then directly executes the action primitive sequence by computing the actions required to reach each point.
We evaluate two versions of MOKA: the first with precise resets that closely match the inputs to the VLM, and the second with slight perturbations to these initial conditions that is representative of environment resets after backward policy rollouts. The comparison seeks to highlight the benefits of closed-loop RL policies over open-loop primitives. 
In Table~\ref{tab:moka}, we see that MOKA with precise resets succeeds between $60\%$ to $100\%$ of the time depending on the task, which actually outperforms KAGI.
However, without these precise resets, the performance of MOKA drops noticeably, as MOKA uses pre-defined motion primitives that are not robust to even slight environment perturbations, since actions are computed directly with respect to specific VLM-generated keypoints on the objects.
However, KAGI can recover most of this performance loss with online fine-tuning using VLM-generated waypoint trajectories.
KAGI leverages the benefits of closed-loop systems that are more generalizable, robust to environment changes, and exhibit retry behavior, as opposed to open-loop systems that use hand-designed action primitives. This makes KAGI more suitable for the autonomous RL paradigm, which involves potentially imperfect policy rollouts on a real robot.


\section{Conclusion \& Future Work} \label{sec:conclusion}
Our experiments present several insights into the opportunities and challenges of autonomous RL pipelines, and the potential of mark-based visual prompting for improving generalization capabilities of robots equipped with RL. In particular, we explore the challenges of using only sparse rewards for online fine-tuning and relying on in-domain demonstrations with low multimodality to pre-train policies for online RL. Approaches using only sparse rewards are slower to learn to complete tasks, and approaches reliant on in-domain demonstrations are much more brittle and less robust to generalizing to new tasks, objects, and environments.

We demonstrate that dense shaping rewards extracted zero-shot from VLMs can help speed up online RL, and facilitate generalization to new tasks where relying only on sparse rewards is less efficient. Leveraging both dense and sparse rewards improves autonomous learning, with better robustness to reduced in-domain demonstrations than sparse rewards alone. Our reward formulation can modify existing fine-tuning methods to make them more robust and data-efficient. Our work demonstrates the benefits of VLM-based dense shaping rewards, and opens up new avenues of exploration to harness the generalization capabilities of VLMs to enhance the robustness of robot learning systems. 

There are two key areas for future work. First, sparse task completion rewards only track if the object reached the target location or not. While this was successful in KAGI, incorporating object-centric waypoint trajectories into dense reward computation via pixel trackers~\cite{karaev2023cotracker} could further expedite learning, especially for longer-horizon tasks. Future work on improving object-centric pixel tracking and occlusion handling would aid this approach. Second, further research into improving offline RL algorithms (CalQL) could improve the performance of base policies that are fine-tuned with dense and sparse rewards, further reducing the number of online fine-tuning steps. Overall, these directions can help efficiently scale autonomous RL systems to more complex tasks with a greater diversity of objects and environments.




\bibliographystyle{IEEEtran}
\bibliography{IEEEabrv,references}

\begin{thebibliography}{10}
\providecommand{\url}[1]{#1}
\csname url@samestyle\endcsname
\providecommand{\newblock}{\relax}
\providecommand{\bibinfo}[2]{#2}
\providecommand{\BIBentrySTDinterwordspacing}{\spaceskip=0pt\relax}
\providecommand{\BIBentryALTinterwordstretchfactor}{4}
\providecommand{\BIBentryALTinterwordspacing}{\spaceskip=\fontdimen2\font plus
\BIBentryALTinterwordstretchfactor\fontdimen3\font minus \fontdimen4\font\relax}
\providecommand{\BIBforeignlanguage}[2]{{%
\expandafter\ifx\csname l@#1\endcsname\relax
\typeout{** WARNING: IEEEtran.bst: No hyphenation pattern has been}%
\typeout{** loaded for the language `#1'. Using the pattern for}%
\typeout{** the default language instead.}%
\else
\language=\csname l@#1\endcsname
\fi
#2}}
\providecommand{\BIBdecl}{\relax}
\BIBdecl

\bibitem{kumar2020conservative}
A.~Kumar, A.~Zhou, G.~Tucker, and S.~Levine, ``Conservative q-learning for offline reinforcement learning,'' \emph{Advances in Neural Information Processing Systems}, vol.~33, pp. 1179--1191, 2020.

\bibitem{ashvin2020accelerating}
A.~Nair, A.~Gupta, M.~Dalal, and S.~Levine, ``Awac: Accelerating online reinforcement learning with offline datasets,'' in \emph{arXiv preprint arXiv:2006.09359}, 2021.

\bibitem{kostrikov2021offline}
I.~Kostrikov, A.~Nair, and S.~Levine, ``Offline reinforcement learning with implicit q-learning,'' \emph{arXiv preprint arXiv:2110.06169}, 2021.

\bibitem{kumar2022pre}
A.~Kumar, A.~Singh, F.~Ebert, M.~Nakamoto, Y.~Yang, C.~Finn, and S.~Levine, ``Pre-training for robots: Offline rl enables learning new tasks in a handful of trials,'' in \emph{Robotics: Science and Systems}, 2022.

\bibitem{mark2022fine}
M.~S. Mark, A.~Ghadirzadeh, X.~Chen, and C.~Finn, ``Fine-tuning offline policies with optimistic action selection,'' in \emph{Deep Reinforcement Learning Workshop NeurIPS 2022}, 2022.

\bibitem{yang2023robofume}
J.~Yang, M.~S. Mark, B.~Vu, A.~Sharma, J.~Bohg, and C.~Finn, ``Robot fine-tuning made easy: Pre-training rewards and policies for autonomous real-world reinforcement learning,'' in \emph{2024 IEEE International Conference on Robotics and Automation}.\hskip 1em plus 0.5em minus 0.4em\relax IEEE, 2024.

\bibitem{ho2016generative}
J.~Ho and S.~Ermon, ``Generative adversarial imitation learning,'' \emph{Advances in Neural Information Processing Systems}, vol.~29, 2016.

\bibitem{fu2017learning}
J.~Fu, K.~Luo, and S.~Levine, ``Learning robust rewards with adverserial inverse reinforcement learnin,'' in \emph{International Conference on Learning Representations}, 2018.

\bibitem{fu2018variational}
J.~Fu, A.~Singh, D.~Ghosh, L.~Yang, and S.~Levine, ``Variational inverse control with events: A general framework for data-driven reward definition,'' \emph{Advances in Neural Information Processing Systems}, vol.~31, 2018.

\bibitem{xie2018few}
A.~Xie, A.~Singh, S.~Levine, and C.~Finn, ``Few-shot goal inference for visuomotor learning and planning,'' in \emph{Conference on Robot Learning}, 2018, pp. 40--52.

\bibitem{singh2019end}
A.~Singh, L.~Yang, K.~Hartikainen, C.~Finn, and S.~Levine, ``End-to-end robotic reinforcement learning without reward engineering,'' in \emph{Robotics: Science and Systems}, 2019.

\bibitem{driess2023palm}
D.~Driess, F.~Xia, M.~S.~M. Sajjadi, C.~Lynch, A.~Chowdhery, B.~Ichter, A.~Wahid, J.~Tompson, Q.~Vuong, T.~Yu, W.~Huang, Y.~Chebotar, P.~Sermanet, D.~Duckworth, S.~Levine, V.~Vanhoucke, K.~Hausman, M.~Toussaint, K.~Greff, A.~Zeng, I.~Mordatch, and P.~Florence, ``Palm-e: An embodied multimodal language model,'' in \emph{arXiv preprint arXiv:2303.03378}, 2023.

\bibitem{brohan2023rt}
A.~Brohan, N.~Brown, J.~Carbajal, Y.~Chebotar, X.~Chen, K.~Choromanski, T.~Ding, D.~Driess, A.~Dubey, C.~Finn \emph{et~al.}, ``Rt-2: Vision-language-action models transfer web knowledge to robotic control,'' \emph{arXiv preprint arXiv:2307.15818}, 2023.

\bibitem{saycan2022arxiv}
M.~Ahn, A.~Brohan, N.~Brown, Y.~Chebotar, O.~Cortes, B.~David, C.~Finn, C.~Fu, K.~Gopalakrishnan, K.~Hausman \emph{et~al.}, ``Do as i can, not as i say: Grounding language in robotic affordances,'' in \emph{Conference on Robot Learning}, 2022.

\bibitem{chen2023open}
B.~Chen, F.~Xia, B.~Ichter, K.~Rao, K.~Gopalakrishnan, M.~S. Ryoo, A.~Stone, and D.~Kappler, ``Open-vocabulary queryable scene representations for real world planning,'' in \emph{IEEE International Conference on Robotics and Automation}.\hskip 1em plus 0.5em minus 0.4em\relax IEEE, 2023, pp. 11\,509--11\,522.

\bibitem{huang2022language}
W.~Huang, P.~Abbeel, D.~Pathak, and I.~Mordatch, ``Language models as zero-shot planners: Extracting actionable knowledge for embodied agents,'' \emph{arXiv preprint arXiv:2201.07207}, 2022.

\bibitem{huang2022inner}
W.~Huang, F.~Xia, T.~Xiao, H.~Chan, J.~Liang, P.~Florence, A.~Zeng, J.~Tompson, I.~Mordatch, Y.~Chebotar, P.~Sermanet, N.~Brown, T.~Jackson, L.~Luu, S.~Levine, K.~Hausman, and B.~Ichter, ``Inner monologue: Embodied reasoning through planning with language models,'' in \emph{Conference on Robot Learning}, 2022.

\bibitem{yang2023set}
J.~Yang, H.~Zhang, F.~Li, X.~Zou, C.~Li, and J.~Gao, ``Set-of-mark prompting unleashes extraordinary visual grounding in gpt-4v,'' \emph{arXiv preprint arXiv:2310.11441}, 2023.

\bibitem{chen2024spatialvlm}
B.~Chen, Z.~Xu, S.~Kirmani, B.~Ichter, D.~Driess, P.~Florence, D.~Sadigh, L.~Guibas, and F.~Xia, ``Spatialvlm: Endowing vision-language models with spatial reasoning capabilities,'' in \emph{Conference on Computer Vision and Pattern Recognition}, 2024.

\bibitem{nasiriany2024pivot}
S.~Nasiriany, F.~Xia, W.~Yu, T.~Xiao, J.~Liang, I.~Dasgupta, A.~Xie, D.~Driess, A.~Wahid, Z.~Xu \emph{et~al.}, ``Pivot: Iterative visual prompting elicits actionable knowledge for vlms,'' \emph{arXiv preprint arXiv:2402.07872}, 2024.

\bibitem{wang2023prompt}
Y.-J. Wang, B.~Zhang, J.~Chen, and K.~Sreenath, ``Prompt a robot to walk with large language models,'' \emph{arXiv preprint arXiv:2309.09969}, 2023.

\bibitem{jiang2022vima}
Y.~Jiang, A.~Gupta, Z.~Zhang, G.~Wang, Y.~Dou, Y.~Chen, L.~Fei-Fei, A.~Anandkumar, Y.~Zhu, and L.~Fan, ``Vima: General robot manipulation with multimodal prompts,'' in \emph{Fortieth International Conference on Machine Learning}, 2023.

\bibitem{liang2023code}
J.~Liang, W.~Huang, F.~Xia, P.~Xu, K.~Hausman, B.~Ichter, P.~Florence, and A.~Zeng, ``Code as policies: Language model programs for embodied control,'' in \emph{arXiv preprint arXiv:2209.07753}, 2022.

\bibitem{huang2023voxposer}
W.~Huang, C.~Wang, R.~Zhang, Y.~Li, J.~Wu, and L.~Fei-Fei, ``Voxposer: Composable 3d value maps for robotic manipulation with language models,'' \emph{arXiv preprint arXiv:2307.05973}, 2023.

\bibitem{ma2023eureka}
Y.~J. Ma, W.~Liang, G.~Wang, D.-A. Huang, O.~Bastani, D.~Jayaraman, Y.~Zhu, L.~Fan, and A.~Anandkumar, ``Eureka: Human-level reward design via coding large language models,'' in \emph{International Conference on Learning Representations}, 2024.

\bibitem{yu2023language}
W.~Yu, N.~Gileadi, C.~Fu, S.~Kirmani, K.-H. Lee, M.~G. Arenas, H.-T.~L. Chiang, T.~Erez, L.~Hasenclever, J.~Humplik, B.~Ichter, T.~Xiao, P.~Xu, A.~Zeng, T.~Zhang, N.~Heess, D.~Sadigh, J.~Tan, Y.~Tassa, and F.~Xia, ``Language to rewards for robotic skill synthesis,'' in \emph{Conference on Robot Learning}, 2023.

\bibitem{shah2022lmnav}
D.~Shah, B.~Osinski, B.~Ichter, and S.~Levine, ``{LM}-nav: Robotic navigation with large pre-trained models of language, vision, and action,'' in \emph{6th Annual Conference on Robot Learning}, 2022.

\bibitem{mahmoudieh2022zeroshotreward}
P.~Mahmoudieh, D.~Pathak, and T.~Darrell, ``Zero-shot reward specification via grounded natural language,'' in \emph{Proceedings of the 39th International Conference on Machine Learning}, vol. 162.\hskip 1em plus 0.5em minus 0.4em\relax PMLR, 2022, pp. 14\,743--14\,752.

\bibitem{villa2023pivot}
A.~Villa, J.~L. Alcázar, M.~Alfarra, K.~Alhamoud, J.~Hurtado, F.~C. Heilbron, A.~Soto, and B.~Ghanem, ``Pivot: Prompting for video continual learning,'' in \emph{2023 IEEE/CVF Conference on Computer Vision and Pattern Recognition}, 2023, pp. 24\,214--24\,223.

\bibitem{bommasani2022opportunities}
R.~B. et~al., ``On the opportunities and risks of foundation models,'' in \emph{arXiv preprint arXiv:2108.07258}, 2022.

\bibitem{zeng2022socraticmodels}
A.~Zeng, M.~Attarian, B.~Ichter, K.~Choromanski, A.~Wong, S.~Welker, F.~Tombari, A.~Purohit, M.~Ryoo, V.~Sindhwani, J.~Lee, V.~Vanhoucke, and P.~Florence, ``Socratic models: Composing zero-shot multimodal reasoning with language,'' in \emph{International Conference on Learning Representations}, 2023.

\bibitem{raman:neurips22fmdm}
S.~S. Raman, V.~Cohen, E.~Rosen, I.~Idrees, D.~Paulius, and S.~Tellex, ``Planning with large language models via corrective re-prompting,'' in \emph{Foundation Models for Decision Making Workshop at NeurIPS 2022}, 2022.

\bibitem{Silver2023GeneralizedPI}
T.~Silver, S.~Dan, K.~Srinivas, J.~B. Tenenbaum, L.~P. Kaelbling, and M.~Katz, ``Generalized planning in pddl domains with pretrained large language models,'' in \emph{AAAI Conference on Artificial Intelligence}, 2023.

\bibitem{LinAgiaEtAl2023}
K.~Lin, C.~Agia, T.~Migimatsu, M.~Pavone, and J.~Bohg, ``Text2motion: from natural language instructions to feasible plans,'' \emph{Autonomous Robots}, Nov 2023.

\bibitem{dorbala2022clipnavusingclipzeroshot}
V.~S. Dorbala, G.~Sigurdsson, R.~Piramuthu, J.~Thomason, and G.~S. Sukhatme, ``Clip-nav: Using clip for zero-shot vision-and-language navigation,'' in \emph{arXiv preprint arXiv:2211.16649}, 2022.

\bibitem{huang23vlmaps}
C.~Huang, O.~Mees, A.~Zeng, and W.~Burgard, ``Visual language maps for robot navigation,'' in \emph{Proceedings of the IEEE International Conference on Robotics and Automation}, London, UK, 2023.

\bibitem{rt22023arxiv}
A.~Brohan, N.~Brown, J.~Carbajal, Y.~Chebotar, X.~Chen, K.~Choromanski, T.~Ding, D.~Driess, A.~Dubey, C.~Finn, P.~Florence, C.~Fu, M.~G. Arenas, K.~Gopalakrishnan, K.~Han, K.~Hausman, A.~Herzog, J.~Hsu, B.~Ichter, A.~Irpan, N.~Joshi, R.~Julian, D.~Kalashnikov, Y.~Kuang, I.~Leal, L.~Lee, T.-W.~E. Lee, S.~Levine, Y.~Lu, H.~Michalewski, I.~Mordatch, K.~Pertsch, K.~Rao, K.~Reymann, M.~Ryoo, G.~Salazar, P.~Sanketi, P.~Sermanet, J.~Singh, A.~Singh, R.~Soricut, H.~Tran, V.~Vanhoucke, Q.~Vuong, A.~Wahid, S.~Welker, P.~Wohlhart, J.~Wu, F.~Xia, T.~Xiao, P.~Xu, S.~Xu, T.~Yu, and B.~Zitkovich, ``Rt-2: Vision-language-action models transfer web knowledge to robotic control,'' in \emph{arXiv preprint arXiv:2307.15818}, 2023.

\bibitem{gu2023rttrajectory}
J.~Gu, S.~Kirmani, P.~Wohlhart, Y.~Lu, M.~G. Arenas, K.~Rao, W.~Yu, C.~Fu, K.~Gopalakrishnan, Z.~Xu, P.~Sundaresan, P.~Xu, H.~Su, K.~Hausman, C.~Finn, Q.~Vuong, and T.~Xiao, ``Rt-trajectory: Robotic task generalization via hindsight trajectory sketches,'' 2023.

\bibitem{3dllm}
Y.~Hong, H.~Zhen, P.~Chen, S.~Zheng, Y.~Du, Z.~Chen, and C.~Gan, ``3d-llm: Injecting the 3d world into large language models,'' in \emph{Advances in Neural Information Processing Systems}, 2023.

\bibitem{huang2023chat}
H.~Huang, Z.~Wang, R.~Huang, L.~Liu, X.~Cheng, Y.~Zhao, T.~Jin, and Z.~Zhao, ``Chat-3d v2: Bridging 3d scene and large language models with object identifiers,'' \emph{arXiv preprint arXiv:2312.08168}, 2023.

\bibitem{wang2023chat}
Z.~Wang, H.~Huang, Y.~Zhao, Z.~Zhang, and Z.~Zhao, ``Chat-3d: Data-efficiently tuning large language model for universal dialogue of 3d scenes,'' \emph{arXiv preprint arXiv:2308.08769}, 2023.

\bibitem{openai2024gpt4ocard}
OpenAI, ``Gpt-4o system card,'' in \emph{arXiv preprint arXiv:2410.21276}, 2024.

\bibitem{geminiteam2024gemini}
G.~T. Google, ``Gemini: A family of highly capable multimodal models,'' in \emph{arXiv preprint arXiv:2312.11805}, 2024.

\bibitem{nair2022r3m}
S.~Nair, A.~Rajeswaran, V.~Kumar, C.~Finn, and A.~Gupta, ``R3m: A universal visual representation for robot manipulation,'' in \emph{Conference on Robot Learning}, 2022.

\bibitem{wang2024rlvlmf}
Y.~Wang, Z.~Sun, J.~Zhang, Z.~Xian, E.~Biyik, D.~Held, and Z.~Erickson, ``Rl-vlm-f: Reinforcement learning from vision language foundation model feedback,'' in \emph{Proceedings of the 41th International Conference on Machine Learning}, 2024.

\bibitem{huang2024rekep}
W.~Huang, C.~Wang, Y.~Li, R.~Zhang, and L.~Fei-Fei, ``Rekep: Spatio-temporal reasoning of relational keypoint constraints for robotic manipulation,'' in \emph{Conference on Robot Learning}, 2024.

\bibitem{ravi2024sam2}
\BIBentryALTinterwordspacing
N.~Ravi, V.~Gabeur, Y.-T. Hu, R.~Hu, C.~Ryali, T.~Ma, H.~Khedr, R.~R{\"a}dle, C.~Rolland, L.~Gustafson, E.~Mintun, J.~Pan, K.~V. Alwala, N.~Carion, C.-Y. Wu, R.~Girshick, P.~Doll{\'a}r, and C.~Feichtenhofer, ``Sam 2: Segment anything in images and videos,'' \emph{arXiv preprint arXiv:2408.00714}, 2024. [Online]. Available: \url{https://arxiv.org/abs/2408.00714}
\BIBentrySTDinterwordspacing

\bibitem{sharma2022autonomous}
A.~Sharma, K.~Xu, N.~Sardana, A.~Gupta, K.~Hausman, S.~Levine, and C.~Finn, ``Autonomous reinforcement learning: Formalism and benchmarking,'' in \emph{International Conference on Learning Representations}, 2022.

\bibitem{balsells2023gear}
M.~Balsells, M.~Torne, Z.~Wang, S.~Desai, P.~Agrawal, and A.~Gupta, ``Autonomous robotic reinforcement learning with asynchronous human feedback,'' in \emph{Conference on Robot Learning}, 2023.

\bibitem{gupta2021resetfree}
A.~Gupta, J.~Yu, T.~Z. Zhao, V.~Kumar, A.~Rovinsky, K.~Xu, T.~Devlin, and S.~Levine, ``Reset-free reinforcement learning via multi-task learning: Learning dexterous manipulation behaviors without human intervention,'' in \emph{2021 IEEE International Conference on Robotics and Automation}, 2021, pp. 6664--6671.

\bibitem{sun2022autonomous}
C.~Sun, J.~Orbik, C.~M. Devin, B.~H. Yang, A.~Gupta, G.~Berseth, and S.~Levine, ``Fully autonomous real-world reinforcement learning with applications to mobile manipulation,'' in \emph{Proceedings of the 5th Conference on Robot Learning}, ser. Proceedings of Machine Learning Research, vol. 164.\hskip 1em plus 0.5em minus 0.4em\relax PMLR, 2022, pp. 308--319.

\bibitem{ebert2021bridge}
F.~Ebert, Y.~Yang, K.~Schmeckpeper, B.~Bucher, G.~Georgakis, K.~Daniilidis, C.~Finn, and S.~Levine, ``Bridge data: Boosting generalization of robotic skills with cross-domain datasets,'' in \emph{Robotics: Science and Systems}, 2022.

\bibitem{walke2023bridgedata}
H.~Walke, K.~Black, A.~Lee, M.~J. Kim, M.~Du, C.~Zheng, T.~Zhao, P.~Hansen-Estruch, Q.~Vuong, A.~He, V.~Myers, K.~Fang, C.~Finn, and S.~Levine, ``Bridgedata v2: A dataset for robot learning at scale,'' in \emph{Conference on Robot Learning}, 2023.

\bibitem{zhu2023minigpt4}
D.~Zhu, J.~Chen, X.~Shen, X.~Li, and M.~Elhoseiny, ``Minigpt-4: Enhancing vision-language understanding with advanced large language models,'' in \emph{International Conference on Learning Representations}, 2024.

\bibitem{xiong2024adaptive}
H.~Xiong, R.~Mendonca, K.~Shaw, and D.~Pathak, ``Adaptive mobile manipulation for articulated objects in the open world,'' \emph{arXiv preprint arXiv:2401.14403}, 2024.

\bibitem{ren2024groundingsam}
T.~Ren, S.~Liu, A.~Zeng, J.~Lin, K.~Li, H.~Cao, J.~Chen, X.~Huang, Y.~Chen, F.~Yan, Z.~Zeng, H.~Zhang, F.~Li, J.~Yang, H.~Li, Q.~Jiang, and L.~Zhang, ``Grounded sam: Assembling open-world models for diverse visual tasks,'' in \emph{arXiv preprint arXiv:2401.14159}, 2024.

\bibitem{karaev2023cotracker}
N.~Karaev, I.~Rocco, B.~Graham, N.~Neverova, A.~Vedaldi, and C.~Rupprecht, ``Cotracker: It is better to track together,'' in \emph{European Conference on Computer Vision}, 2024.

\bibitem{sharma2023medal++}
A.~Sharma, A.~M. Ahmed, R.~Ahmad, and C.~Finn, ``Self-improving robots: End-to-end autonomous visuomotor reinforcement learning,'' in \emph{7th Annual Conference on Robot Learning}, 2023.

\bibitem{nakamoto2023calql}
M.~Nakamoto, Y.~Zhai, A.~Singh, M.~S. Mark, Y.~Ma, C.~Finn, A.~Kumar, and S.~Levine, ``Cal-ql: Calibrated offline rl pre-training for efficient online fine-tuning,'' in \emph{Advances in Neural Information Processing Systems}, 2023.

\bibitem{liu2024moka}
K.~Fang, F.~Liu, P.~Abbeel, and S.~Levine, ``Moka: Open-world robotic manipulation through mark-based visual prompting,'' in \emph{Robotics: Science and Systems}, 2024.

\end{thebibliography}

\section*{Appendix}
\textbf{Verifying Reproduction of RoboFuME.} Since our proposed dense shaping rewards to improve online policy fine-tuning are integrated into RoboFuME~\cite{yang2023robofume}, we performed preliminary experiments on two tasks featured in~\cite{yang2023robofume} -- Cloth Folding and Cloth Covering -- and verified their successful reproduction. We also performed additional experiments verifying RoboFuME's reliance on high-quality in-domain demonstrations with low multimodality for both pre-training RL policies and fine-tuning the sparse reward classifier, showing that eliminating in-domain demonstrations entirely was catastrophic for the system's performance. Finally, we explore RoboFuME's selection of Bridge data subsets~\cite{ebert2021bridge, walke2023bridgedata} for pre-training their offline RL policies for tabletop manipulation, which we use in our experiments. 

Overall, our preliminary experiments revealed that there are robustness challenges for pipelines relying on a large number of high-quality in-domain demonstrations, and incorporating task-relevant offline data facilitates transfer to the desired task. These key findings motivated the methodological details of KAGI for improving the efficiency of online fine-tuning with VLM-generated dense shaping rewards.

\textbf{Qualitative Analysis of VLM Performance.} The performance of our online RL fine-tuning system is influenced by the accuracy of VLM keypoint and waypoint predictions. While MOKA verifies that keypoint-based reasoning of VLMs are reliably accurate for tabletop manipulation and generalizable to new objects and tasks, the paper only provides a single top-down viewpoint to the VLM, since MOKA's open-loop primitives compute actions based on pre-determined depth information specific to the evaluation setup, which requires manual engineering and tuning.

We observe that for closed-loop RL systems, providing dual-angle viewpoints are important to accurately generate waypoint trajectories in 3D space for dense rewards. Furthermore, providing in-context examples help facilitate waypoint validity. We modify MOKA's metaprompt to accommodate this additional dimension of input, so as to generate 3D waypoints instead of 2D waypoints for KAGI's dense rewards. We keep this metaprompt the same across all tasks. To test the robustness of our prompting structure, in addition to GPT-4o, we analyze the outputs generated from our prompts when provided to GPT-4V (used by MOKA, larger but slower than GPT-4o) and Gemini 1.5. Qualitative analysis of each VLM's outputs on our tasks determines that Gemini performs similarly and can replace GPT-4o in our framework. GPT-4V, while generally reliable, can struggle to generate sensible depth movements using the side-angle view to facilitate successful task completion. Overall, KAGI is designed to flexibly incorporate VLMs of similar or better reasoning capability compared to GPT-4o, and we leave extensive qualitative comparisons and benchmarking frameworks for VLMs on 3D spatial reasoning tasks to future work.

Please see the Additional Information section of our \href{https://sites.google.com/view/affordance-guided-rl}{project page} for more details on these areas of analysis and other preliminary experiments.

\end{document}